\documentclass[sigplan,screen]{acmart}

\setcopyright{none}
\renewcommand\footnotetextcopyrightpermission[1]{}
\usepackage{color}
\usepackage{xcolor}
\usepackage{bm}
\usepackage{graphicx}
\usepackage{subfigure}
\usepackage{caption}

\usepackage{epstopdf}
\usepackage{multirow}
\usepackage{algorithm}
\usepackage{algorithmic}
\usepackage{mathpazo}
\usepackage{xspace}

\AtBeginDocument{%
  \providecommand\BibTeX{{%
    \normalfont B\kern-0.5em{\scshape i\kern-0.25em b}\kern-0.8em\TeX}}}

\begin{document}
\bibliographystyle{unsrt} 
\title{Exploration of Input Patterns for Enhancing the Performance of Liquid State Machines}

\author{Shasha Guo}
\affiliation{%
  \institution{National University of Defense Technology}
  \city{Changsha}
  \country{China}}
\email{guoshasha13@nudt.edu.cn}

\author{Lianhua Qu}
\affiliation{%
  \institution{National University of Defense Technology}
  \city{Changsha}
  \country{China}}

\author{Lei Wang}
\affiliation{%
  \institution{National University of Defense Technology}
  \city{Changsha}
  \country{China}}
\email{arrowya@gmail.com}

\author{Shuo Tian}
\affiliation{%
  \institution{National University of Defense Technology}
  \city{Changsha}
  \country{China}}

\author{Shiming Li}
\affiliation{%
  \institution{National University of Defense Technology}
  \city{Changsha}
  \country{China}}

\author{Weixia Xu}
\affiliation{%
  \institution{National University of Defense Technology}
  \city{Changsha}
  \country{China}}
\begin{abstract}
  Spiking Neural Networks (SNN) have gain increasing attention for its low power consumption. But training SNN is challenging. Liquid State Machine (LSM), as a major type of Reservoir computing, has been widely recognized for its low training cost among SNNs.
    The exploration on LSM topology for enhancing performance often requires hyper-parameter search, which is both resource expensive and time consuming. We explore the influence of input scale reduction on LSM instead. There are two main reasons for studying input reduction of LSM. One is that input dimension of large images requires efficient processing. Another one is that input exploration is generally more economic than architecture search. To mitigate the difficulty in effectively dealing with huge input spaces of LSM, and to find that whether input reduction can enhance LSM performance, we explore serveral input patterns, namely fullscale, scanline, chessboard and patch. Several datasets have been used to evaluate the performance of the proposed input patterns, including two spatio image datasets and one spatio-temporal image database. The experimental results show that the reduced input under chessboard pattern improves the accuracy by up to 2\%, and reduces execution time by up to 50\% with up to 75\% less input storage than the fullscale input pattern for LSM.
\end{abstract}



\keywords{LSM, encoding, input pattern}


\maketitle

\section{Introduction}

In the past decades, both academia and industry have made great attempts to develop computing models to mimic brain function. 
While spiking neural networks (SNNs) hold a lot of promise due to its closer resemblance to brains than older generations of artificial neural networks and are anticipated to be power efficient, the training of SNNs is nontrivial. There are several unsupervised learning algorithms, such as Winner-Take-All strategy \cite{li2012selective} and spike-timing-dependent plasticity \cite{masquelier2007unsupervised}. To develop supervised gradient-based training for SNNs, particularly recurrent SNNs, requires approximation to the neuron model and will consume large amounts of resources and training efforts.

There has been increasing interest in the concept of reservoir computing (RC). The liquid state machine (LSM) is one specific form of RC.
LSM, first proposed by \cite{maass2002real}, has gained increasing popularity due to its lower training cost than other SNNs. 
It has been used for various applications such as image recognition \cite{grzyb2009facial,wang2016d}, movement prediction \cite{maass2002new, burgsteiner2007movement, kaiser2017scaling}, decoding actual brain activity \cite{nikolic2009distributed, al2018anytime}, and speech recognition \cite{zhang2015digital,jin2017performance, zhang2019information}.

To find a LSM with the state-of-the-art results in real-world applications, many efforts of enhancing LSM focus on the exploration of LSM topology, such as self-organizing networks (SON) \cite{luo2018improving}, small-world networks \cite{hazan2012topological}, deep LSM \cite{wang2016d}, and liquid ensembles \cite{wijesinghe2019analysis}. These works either incur training algorithms or cost-intensive hyper-parameter search, which increase accuracy at the cost of performance or resource overhead.

There exist several approaches that could improve the accuracy of LSM without deviating from its inherent structure simplicity, like increasing the number of neurons in the liquid. However, this approach also demands more computations and storage. And the sensitivity of accuracy to neuron count decreases when the number of neurons exceeding a certain point \cite{wijesinghe2019analysis}.

Dealing with input is another issue that has significant impact on the performance of LSM.
Some previous studies focus on the encoding methods of SNN input \cite{xu2017spike, petro2019selection}. 
\cite{xu2017spike} explores the encoding for rate coding of image input. \cite{petro2019selection} compares four temporal encoding method for converting analog values into spikes for SNN input. These two works focus on the encoding or converting method of analog real-values without considering the whole input size. With the increase in the size of images, the converted input size is also growing rapidly to a large scale.
\cite{lin2018programming} uses a new method to encode images for SNN input. It utilizes the redundancy in images and sparsely encodes the images into spikes through certain ``scanlines".

In this work, we aim to systematically explore the effect of reducing the input scale via input pattern on the performance of LSM, that is, accuracy, runtime and storage. 
There are two main reasons for this. First, the number of input pixels represents the number of input layer neurons. Large input processing is more time-consuming and resource-demanding than small input considering the same liquid architecture and synapse connection probability.
Second, LSM performance not only refers to the accuracy, but also the storage and runtime. While previous work of parameter search mainly focus on the accuracy improvement, input pattern search takes consideration of these three aspects.



Our contributions of this paper are as follows:
\begin{itemize}
  \item We explore several input patterns for LSM simulation, and compare their performance in accuracy, execution time and input storage cost. We use three datasets, including two frame-based datasets and one spatio-temporal event-based dataset.
  \item We propose two algorithms for using different input patterns to generate desired spike trains from frame-based datasets and event-based database.
  \item We use several readout layers to show that the effects brought by input patterns are reliable, i.e., the performance change of LSM is indeed brought by the input pattern.

\end{itemize}
We validate our idea on several benchmarks including two frame-based datasets and one event-based neuromorphic dataset. The frame-based datasets are converted to spiking format as input of LSM.
The experimental results show that the reduced input under chessboard pattern can improve the accuracy by up to 2\%, and reduces the execution time by up to 50\% with up to 75\% less input storage than fullscale pattern for LSM.

\section{Preliminaries}
\subsection{LSM}
\begin{figure}[b]
  \includegraphics[width=.3\textwidth]{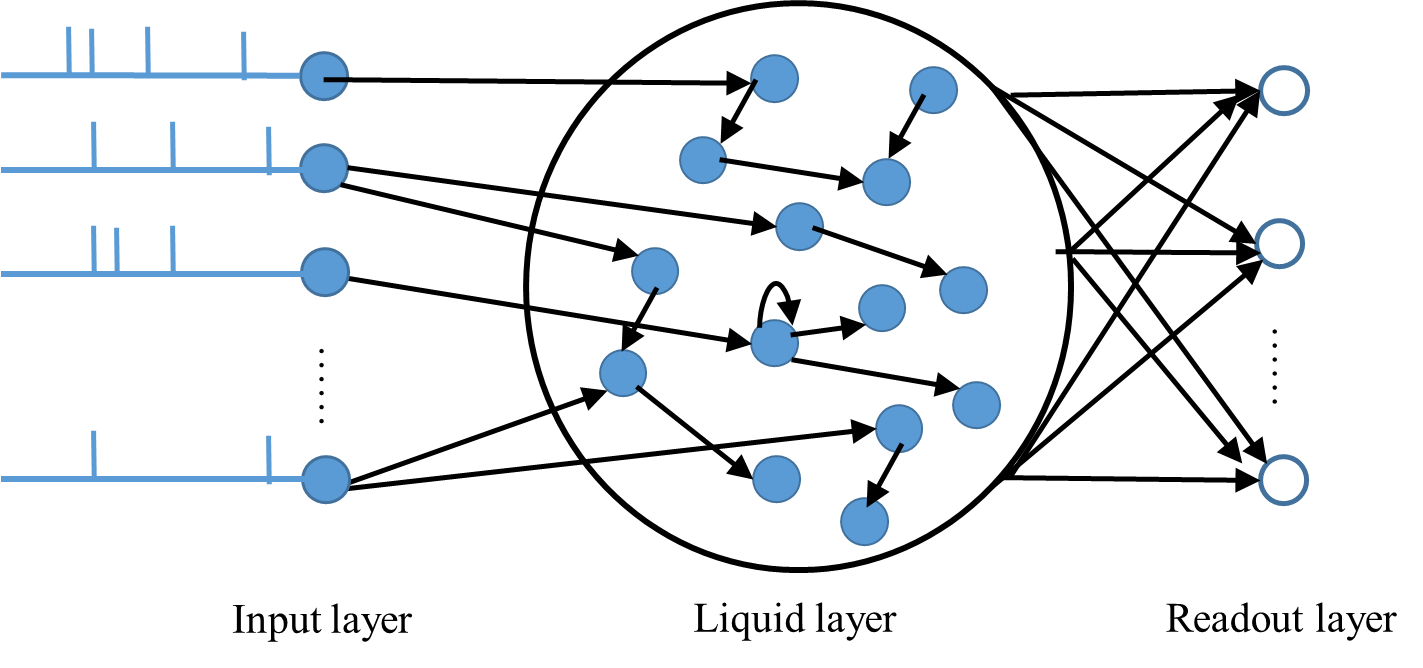}
  \caption{Illustration of LSM.}
  \label{fig:lsm}
\end{figure}

A liquid state machine (as shown in Figure~\ref{fig:lsm}) consists of three main components: an input layer, a liquid layer and a readout layer. Input is fed through the input layer into the liquid. The liquid receives the input streams and transforms them into non-linear patterns in higher dimensions, which acts as a filter. The state vector of the liquid is then used for analysis and interpretation by the readout layer, which consists of either memroyless artificial neurons or spiking neurons.

A liquid neuron can be either excitatory (E) or inhibitory (I) according how it affects other neurons. 
Neurons are connected by synapses, and the weight on a synapse represents how much the pre-synaptic neuron's activity affects the post-synaptic neuron.
The connections between two neurons in the liquids are arranged with probability C. C depends on the type of pre-synaptic and the post-synaptic neurons, that is whether they are the excitatory or inhibitory cells. 
\subsection{Neuron Model}
In our experiments, we use the standard leaky-integrate-and-fire (LIF) neuron model \cite{gerstner2002spiking} to mimic the dynamics of both excitatory and inhibitory spiking neurons. The LIF model is probably one of the most popular spiking neuron models, more biologically realistic than the IF model \cite{koch2004biophysics}. A LIF neuron fires a spike when the membrane potential ($v$) reaches the threshold value ($v_{th}$). After spiking, the membrane potential is reset to the resting value.
The dynamics of the membrane potential of the LIF neuron can be described as follows: $\tau _{m}\frac{dv}{dt}= -v\left ( t \right ) + RI\left ( t \right )$. $I(t)$ is input current, $\tau _{m}$ is the membrane time-constant, and $R$ is the membrane resistance.
The formulations elaborated in \cite{diehl2015unsupervised} are used for modeling the dynamics of the LIF neurons in this paper.

\subsection{Image to Spike Conversion}
In general, SNNs receive and process event-driven spike signals. Thus, the conversion to spike trains is needed when testing SNNs over frame-based datasets in real value.
One of the prevalent strategies is the probabilistic sampling. At every time step, it samples the original pixel intensity (usually normalized to [0, 1]) into a binary value, wherein the probability of being 1 (firing a spike) equals to the intensity value. The sampling follows a given probability distribution such as Bernoulli distribution or Poisson distribution.

\subsection{Training Readout Layer}
We use two algorithms for readout layer training, back propagation algorithm (BP) \cite{rumelhart1986learning} with stochastic gradient descent (SGD), and state vector machine (SVM) \cite{cortes1995support}. The reason of using several classifiers is that we want to explore whether certain input pattern can bring similar effects despite using different readouts.

The readout layer takes the normalized liquid state vectors as input.
The liquid state vectors record the firing number of each neuron for all input images during the simulation, which are then normalized by the most spiking numbers in order to range between 0 and 1.

When using SVM as the readout layer, we add some aggressive operations on the normalized state vectors before inputted into the SVM via a binarization function. If the element exceeds 0.5, the value will be set as 1. Otherwise, it will be set to 0. This is SVM1. The SVM without the binarization step is called SVM2.

\section{Method}
To reduce the difficulty in effectively processing huge input spaces of LSM, we explore the influence of dimension reduction in input layer on LSM performance.

\begin{figure}
\centering
\subfigure[]{
\label{fig:scanline}
  \includegraphics[width=.7in,height=.7in]{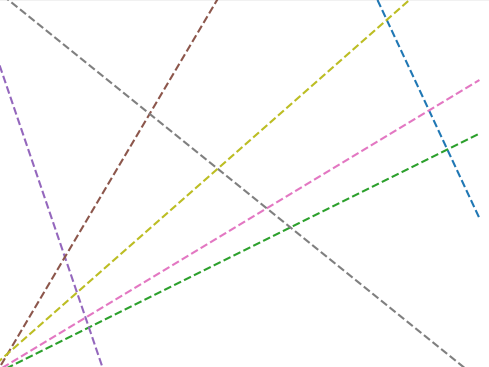}}
\subfigure[]{
\label{fig:chessboard}
  \includegraphics[width=.7in,height=.7in]{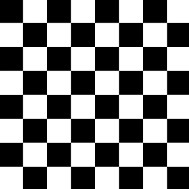}}
\subfigure[]{
\label{fig:patch}
  \includegraphics[width=.7in,height=.7in]{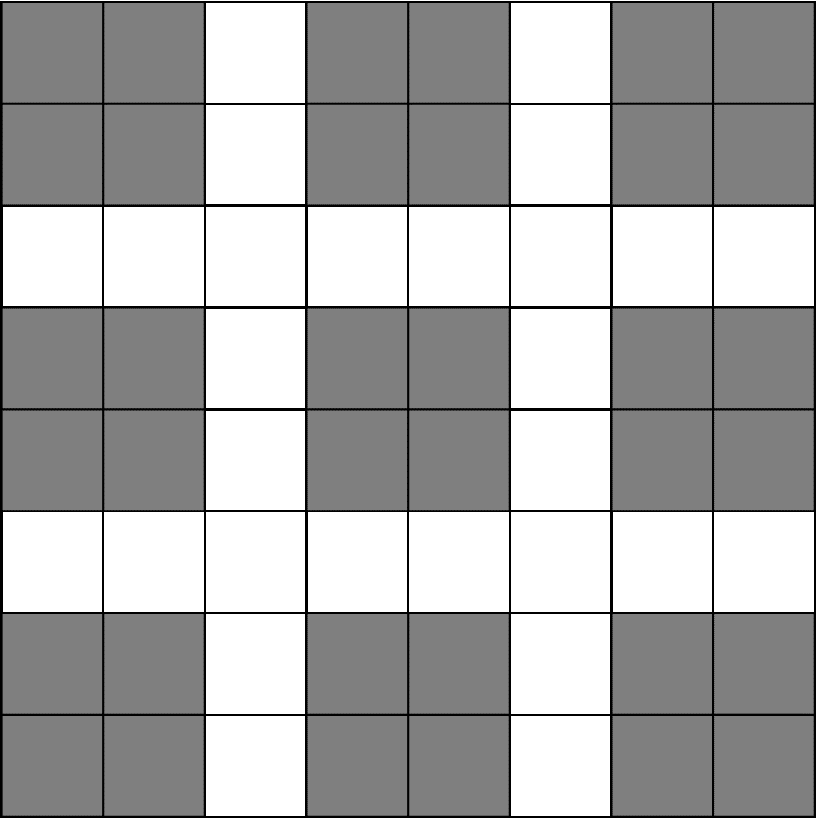}}
\caption{Example of different input patterns. (a) Scanline. (b) Chessboard. (c) Patch.}
\label{fig:input pattern}
\end{figure}

We explore four input patterns, namely, fullscale, scanline, chessboard, and patch.
Fullscale pattern is the baseline, i.e., using all pixels of an image as input. Scanline pattern is used in \cite{lin2018programming} (shown in Figure~\ref{fig:scanline}) as an encoder for images. It is inspired by human saccadic eye movements.

\begin{figure}
\centering
\subfigure[]{
\label{fig:template}
  \includegraphics[width=.3\linewidth]{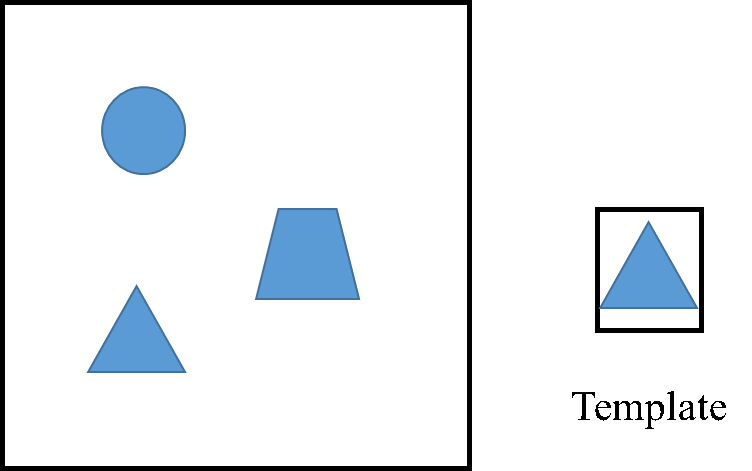}}
  \hspace{0.5in}
\subfigure[]{
\label{fig:keypoint}
  \includegraphics[width=.3\linewidth]{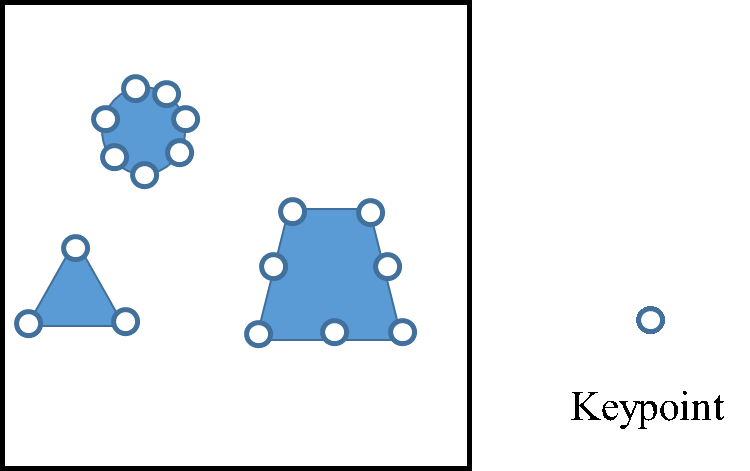}}
\caption{Techniques for feature detection and matching. (a) Template matching. (b) Keypoint detection.}
\label{fig:feature detection}
\end{figure}

The last two input patterns, chessboard and patch, are inspired by two conventional computer vision techniques: template matching and keypoint detection for feature detection and matching. Examples of the two techniques are shown in Figure~\ref{fig:feature detection}. They were used to solve computer vision tasks before the era of artificial neural networks. This reveals that some parts of an image are able to show the significant features for recognition and classification. To reduce the computation cost, we select the pixels and small patches evenly, rather than the way that first computes the importance of the pixels or patches based on some algorithms and then chooses them according to the importance. As shown in Figure~\ref{fig:chessboard} and Figure~\ref{fig:patch}, our approach requires little computing to see whether such simple patterns could have good performance on LSM.

\vspace{2mm}

\vspace{2mm}

 \begin{algorithm}
 \caption{Algorithm for frame-based dataset}
 \label{alg:alg1}
 \footnotesize{
 \begin{algorithmic}[1]
 \renewcommand{\algorithmicrequire}{\bf{Input:}}
 \renewcommand{\algorithmicensure}{\bf{Output:}}
 \REQUIRE input pattern $p$ \\($p$ $\in$ $\{fullscale, scanline, chessboard, patch\}$ ), \\ simulation time:$T$, desired spike train number:$N_{1}$, target set:$X$ (training or testing), number of samples in $X$:$N_{2}$, image height:$H$, image width:$W$
 \ENSURE  spike train index list:$SL$, spike train time list:$TL$
  \STATE generate fixed pixel id list $L$ based on the input pattern and the size of images in $X$
  \STATE initialize a monitor $M$ in BRIAN to record the neuron index and time when a spike occurs
  \FOR {$i = 0$ to $N_{1}$}
  \STATE generate an empty list $R$ as rate list
  \STATE get an image matrix from X indexed by ($i\%N_{2}$)
  \FOR {$j$ in $L$}
  \STATE get the activation value $a$ of pixel $j$ in the image
  \STATE put $a$ into rate list $R$
  \ENDFOR
  \STATE generate Poisson spike trains based on the rate list $R$ in BRIAN
  \ENDFOR
  \STATE get $SL$ and $TL$ from $M$
 \RETURN $SL$ and $TL$
 \end{algorithmic}
 }
 \end{algorithm}
\vspace{-2mm}
\begin{algorithm}
\caption{Algorithm for event-based dataset}
\label{alg:alg2}
\footnotesize{
\begin{algorithmic}[1]
\renewcommand{\algorithmicrequire}{\bf{Input:}}
\renewcommand{\algorithmicensure}{\bf{Output:}}
\REQUIRE input pattern $p$ \\($p$ $\in$ $\{fullscale, scanline, chessboard, patch\}$ ), \\ simulation time:$T$, target set:$X$ (training or testing), image height:$H$, image width:$W$
\ENSURE  spike train index list:$SL$, spike train time list:$TL$
 \STATE generate fixed pixel id list $L$ based on input pattern $p$, $H$ and $W$.
 \STATE sample spike records in X within time T and get four lists $P_{ON}$, $P_{OFF}$, $T_{ON}$ and $T_{OFF}$.
 \STATE $N = H \times W$
 \STATE generate two empty lists $DL_{ON}$ and $DL_{OFF}$ as indices list to be deleted from the above lists.
 \FOR {$i = 0$ to $N$}
 \IF {($i \notin L$)}
 \STATE find the indices at which the $P_{ON}$ equals to $i$ and put these indices into $DL_{ON}$;
 \STATE update $DL_{OFF}$ as above.
 \ENDIF
 \ENDFOR
 \STATE delete elements indexed by $DL_{ON}$ from $P_{ON}$ and $T_{ON}$; delete elements indexed by $DL_{OFF}$ from $P_{OFF}$ and $T_{OFF}$
\RETURN $P_{ON}$, $T_{ON}$, $P_{OFF}$ and $T_{OFF}$
\end{algorithmic}
}
\end{algorithm}
\vspace{-2mm}

Given that there are four input patterns, we need a universal method to generate corresponding spike trains when applying different input patterns on the same dataset.
As frame-based databases and event-based dataset store data in different formats, i.e., images and spike trains, we propose two algorithms, algorithm \ref{alg:alg1} and algorithm \ref{alg:alg2}, to get the desired spike trains for frame-based datasets and event-based dataset respectively. {\bf The algorithms are applicable for all input patterns}.
\section{Experiment Setup}
We use two LSM architectures and three datasets in this work. The liquid was modeled in BRIAN2 \cite{stimberg2019brian}, a Python-based spiking neural network simulator. 

The experimental settings of the two architecture on the three datasets are listed in Table~\ref{tab:experiments}.

\subsection{Benchmark Datasets}
MNIST \cite{lecun1998gradient} dataset is commonly used in machine learning applications in computer vision.
MNIST has 60000 images for training and 10000 other images for testing. Each digit frame is a $28 \times 28$ grayscale image.

\begin{figure}
  \centering
  \includegraphics[width=.3\linewidth]{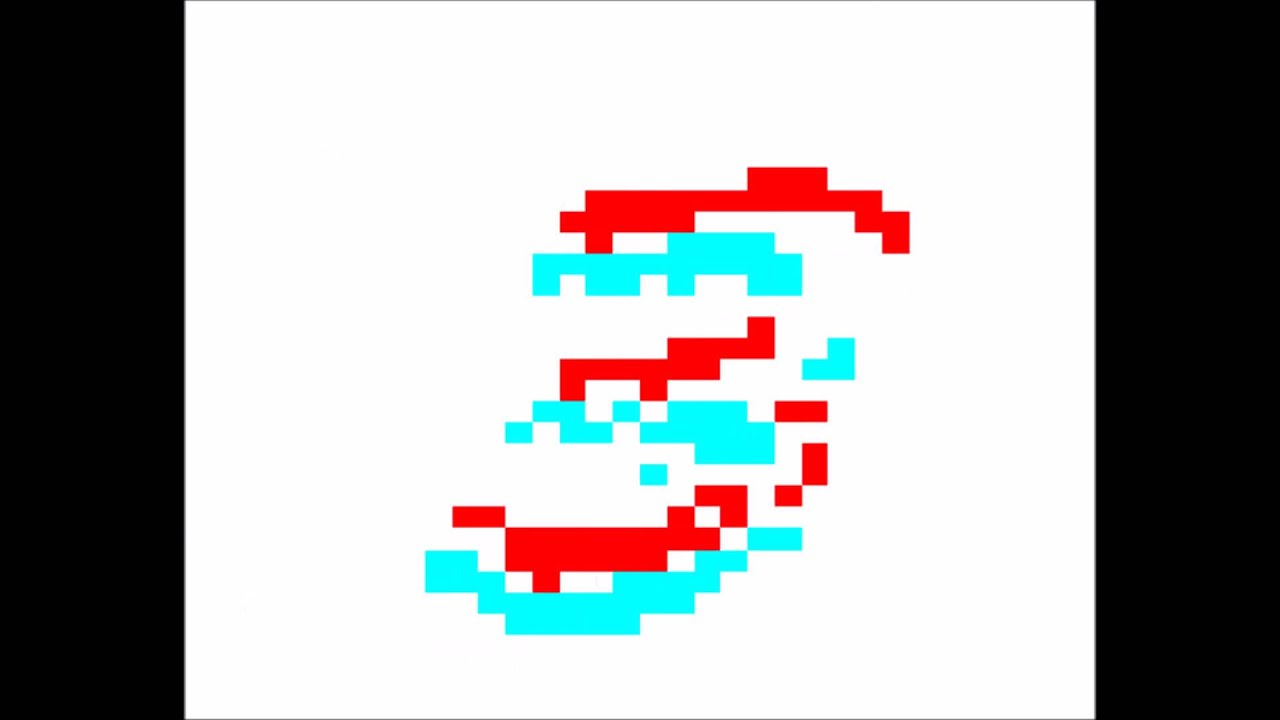}
  \caption{Example of N-MNIST dataset \cite{orchard2015converting}. The two colors represent two event types: ON and OFF.}
  \label{fig:nmnist}
\end{figure}

N-MNIST \cite{orchard2015converting} is the spike version of the orginal MNIST dataset, which also contains 60,000 training samples and 10 000 testing samples.
Each sample in N-MNIST is a spatio-temporal spike pattern with size of $34 \times 34 \times 2 \times T$, where T is the recording time length. An example of N-MNIST is shown in Figure~\ref{fig:nmnist}. {\bf We use a subset of MNIST and NMNIST for experiments.}

\vspace{2mm}

\begin{figure}
\centering
\begin{minipage}[b]{\linewidth}
\includegraphics[width=.45in,height=.45in]{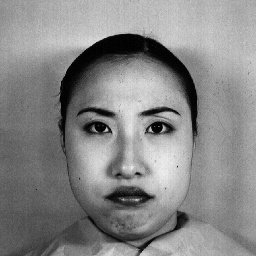}\hspace{-.3pt}
\includegraphics[width=.45in,height=.45in]{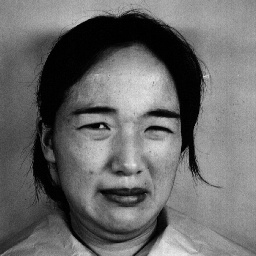}\hspace{-.3pt}
\includegraphics[width=.45in,height=.45in]{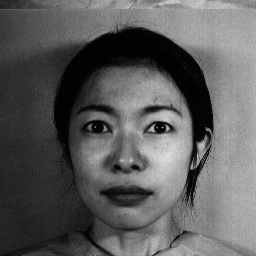}\hspace{-.3pt}
\includegraphics[width=.45in,height=.45in]{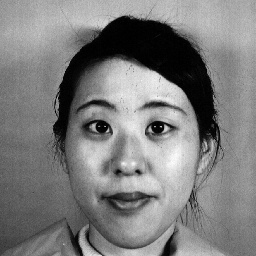}\hspace{-.3pt}
\includegraphics[width=.45in,height=.45in]{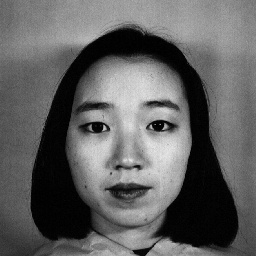}\hspace{-.3pt}
\includegraphics[width=.45in,height=.45in]{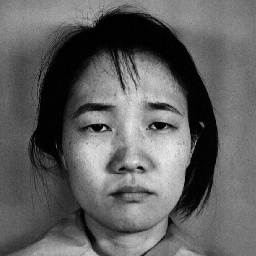}\hspace{-.3pt}
\includegraphics[width=.45in,height=.45in]{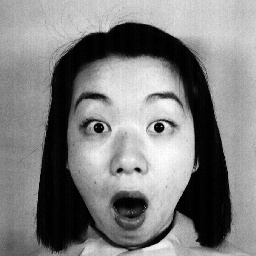}
\end{minipage}

\begin{minipage}[b]{\linewidth}
\includegraphics[width=.15in,height=.45in]{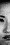}\hspace{-.3pt}
\includegraphics[width=.15in,height=.45in]{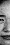}\hspace{-.3pt}
\includegraphics[width=.15in,height=.45in]{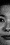}\hspace{-.3pt}
\includegraphics[width=.15in,height=.45in]{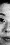}\hspace{-.3pt}
\includegraphics[width=.15in,height=.45in]{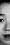}\hspace{-.3pt}
\includegraphics[width=.15in,height=.45in]{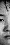}\hspace{-.3pt}
\includegraphics[width=.15in,height=.45in]{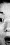}
\end{minipage}

\begin{minipage}[b]{\linewidth}
\includegraphics[width=.45in,height=.45in]{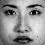}\hspace{-.3pt}
\includegraphics[width=.45in,height=.45in]{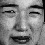}\hspace{-.3pt}
\includegraphics[width=.45in,height=.45in]{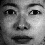}\hspace{-.3pt}
\includegraphics[width=.45in,height=.45in]{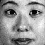}\hspace{-.3pt}
\includegraphics[width=.45in,height=.45in]{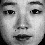}\hspace{-.3pt}
\includegraphics[width=.45in,height=.45in]{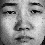}\hspace{-.3pt}
\includegraphics[width=.45in,height=.45in]{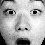}
\end{minipage}
\caption{Examples of JAFFE dataset under different representations.
The upper line shows the original grayscale images with $256 \times 256$ pixels, the middle line shows the representation with $10 \times 45$ pixels and the bottom line shows the other representation with $45 \times 45$ pixels. The facial expression from left to right is anger, disgust, fear, happy, neutral, sadness and surprise.}
\label{fig:jaffe}
\end{figure}

We also make use of the JAFFE \cite{lyons1998coding} dataset, which contains 213 images of female facial expressions posed by 10 Japanese women, grouped into 7 classes.
We use two image representations of JAFFE, by resizing and cropping the images to re-center the eyes, nose, and mouth appear in the picture. One representation contains a small part of the face ($10 \times 45$ pixels) as same as the the size in \cite{grzyb2009facial}. The other representation contains 45 $\times$ 45 pixels. Examples are presented in Figure~\ref{fig:jaffe}.

\subsection{Architectures}

We use two LSM architectures here. One architecture contains a single liquid, which is called 1RC for simplicity. Another architecture is denoted as 5RC, which consists of five liquids in parallel like in \cite{wijesinghe2019analysis}. The total number of neurons in the five liquids will be as same as that in the 1RC for a fair comparison.

\begin{table}
  \centering
  \caption{Experimental Settings.}
  \footnotesize{
    \begin{tabular}[width=\linewidth]{p{4.565em}lllll}
      \toprule
    \multicolumn{1}{l}{Dataset} & Pattern & Arch & \multicolumn{1}{p{2.265em}}{Num} & \multicolumn{1}{p{3.335em}}{Train \newline{}Samples} & \multicolumn{1}{p{3.335em}}{Test \newline{}Samples} \\
    \midrule
    \multicolumn{1}{l}{MNIST}  & All & 1RC, 5RC & 1000  & 10000 & 10000 \\
    \multicolumn{1}{l}{N-MNIST} & FS, CB & 1RC, 5RC & 1000  & 10000 & 10000 \\
    \multicolumn{1}{l}{JAFFE1$^{\mathrm{a}}$} & FS, SL, CB & 1RC & 450 & 1000  & 100 \\
    \multicolumn{1}{l}{JAFFE2$^{\mathrm{b}}$} & FS, SL, CB & 1RC & 1350  & 1000  & 100 \\
    \bottomrule
    \end{tabular}%
  \label{tab:experiments}%
  \\\vspace{1mm}\parbox{1.\linewidth}{$^{\mathrm{a}}$JAFFE1 refers to the representation of 10x45 pixels. \newline{}$^{\mathrm{b}}$JAFFE2 refers to the representation of 45x45 pixels. \newline{}FS is fullscale, SL is scanline, CB is chessboard and PA means Patch. The column {\it Num} means the total number of neurons in the reservoir. }
  }
\end{table}%



All synapse weights, i.e., input-to-liquid, liquid-to-liquid, and liquid-to-readout weights, are initialized from a Gaussian distribution of mean 0.5 and variance 0.16. All the hyper-parameters are listed in Table~\ref{tab:hyperparameter}.

\begin{table}
\caption{Hyper-parameters Value.}
\vspace{2mm}
\footnotesize{
    \begin{tabular}{lccccccc}
      \toprule
          & \multicolumn{1}{c}{$C_{EE}$} & \multicolumn{1}{c}{$C_{EI}$} & \multicolumn{1}{c}{$C_{IE}$} & \multicolumn{1}{c}{$C_{II}$} & IR    & \multicolumn{1}{c}{OR} & \multicolumn{1}{c}{EIR} \\
          \midrule
    (N-)MNIST & 0.4   & 0.4   & 0.5   & 0     & \multicolumn{1}{c}{0.2} & \multirow{2}[0]{*}{0.9} & \multirow{2}[0]{*}{0.8} \\
    JAFFE1/2 & 0.3   & 0.2   & 0.4   & 0.1   & 0.1/0.05 &       &  \\
    \bottomrule
    \end{tabular}%
  \label{tab:hyperparameter}%
\\\vspace{1mm}\parbox{8.3cm}{(N-)MNIST represents MNIST and N-MNIST. $C_{EE}$ is the connection probability of any two excitatory neurons in the liquid, and $C_{EI}$ is the connection probability from an excitatory neuron to an inhibitory neuron. EIR represents the percentage of excitatory neurons in the liquid, IR is the ratio of connections between input layer neurons and the excitatory liquid neurons, OR is the ratio of connections between excitatory liquid neurons and readout layer neurons.}
}
\end{table}
\vspace{1mm}


\section{Experiment results}
The fullscale pattern is regarded as baseline. We used a more aggressive chessboard pattern than Figure~\ref{fig:chessboard} shows. It only samples 1/4 of the fullscale pixels.
\subsection{Accuracy}
We repeat the simulation and training process several times and only the best results are listed for further discussions of MNIST and N-MNIST. For JAFFE we adopt other mechanisms (introduced in section \ref{sec:jaffe}).
\subsubsection{MNIST}
Table~\ref{tab:mnist} reveals the performance of LSM on MNIST dataset using four different input patterns and three classifiers. For training, fullscale pattern always yields the best accuracy. For testing, it can be seen that for 1RC, chessboard pattern brings the best testing accuracy, 0.5\% higher than fullscale pattern. And for 5RC, the difference between fullscale and chessboard pattern are less than 0.3\%. Other two patterns are about 2\% to 4\% worse than fullscale and chessboard.

In addition, we can see that for all readout methods, the accuracy values of chessboard pattern are similar with those of fullscale pattern and over-perform other two patterns. This shows that the effects brought by input patterns are reliable.
And as the SGD method generally performs better than SVM methods, we decide to use SGD classifier for further investigations.

\begin{table}[!htb]
  \centering
  \caption{Accuracy Comparison on MNIST Dataset.}
  \footnotesize{
    \begin{tabular}[width=\linewidth]{p{1.565em}lllllll}
      \toprule
          &       & \multicolumn{3}{c}{1RC} & \multicolumn{3}{c}{5RC} \\
          &       & SGD   & SVM1  & SVM2  & SGD   & SVM1  & SVM2 \\
          \midrule
    \multirow{4}[0]{*}{Train} & FS & \bf{93.90\%} & 74.01\% & 86.31\% & \bf{94.09\%} & 92.39\% & 90.91\% \\
          & SL & 90.64\% & 69.13\% & 81.17\% & 89.42\% & 90.42\% & 86.81\% \\
          & CB & \bf{92.30\%} & 72.04\% & 84.46\% & \bf{93.06\%} & 91.57\% & 89.69\% \\
          & PA & 89.93\% & 70.73\% & 80.65\% & 89.34\% & 88.02\% & 85.48\% \\
          \midrule
    \multirow{4}[0]{*}{Test} & FS & \bf{88.73\%} & 72.24\% & 85.28\% & \bf{89.62\%} & 89.52\% & 89.56\%
    \\
          & SL & 86.37\% & 66.79\% & 79.93\% & 85.39\% & 86.79\% & 85.52\% \\
          & CB & \bf{89.39\%} & 71.61\% & 84.36\% & {\bf 89.35}\% & 89.34\% & 89.27\% \\
          & PA & 86.58\% & 68.58\% & 79.61\% & 86.51\% & 85.15\% & 85.14\% \\
          \bottomrule
    \end{tabular}%
    \\\vspace{1mm}\parbox{8.3cm}{FS is fullscale, SL is scanline, CB is chessboard and PA means Patch.}
    }
  \label{tab:mnist}%
\end{table}%

\subsubsection{N-MNIST}
We train LSM on both channels, i.e., ON and OFF. Table~\ref{tab:nmnist} shows the accuracy comparison of LSM on N-MNIST dataset using SGD for readout layer training. The nearly 10\% differences between training and testing set show that this classifier is over-fitting. It's worth noticing that chessboard pattern is beneficial to mitigate this effect. This is reasonable, as less input is suggested to avoid over-fitting. For inference, with 1RC architecture, using chessboard pattern achieves nearly 5\% higher accuracy than using fullscale pattern. With 5RC architecture, the difference between the two patterns is not significant, but still, chessboard pattern is 1.65\% better than fullscale pattern.
\begin{table}
\caption{Accuracy Comparison on N-MNIST Dataset.}
\vspace{2mm}
\footnotesize{
    \begin{tabular}[width=\linewidth]{lrrrr}
      \toprule
          & \multicolumn{2}{c}{Train} & \multicolumn{2}{c}{Test} \\
          & \multicolumn{1}{l}{fullscale} & \multicolumn{1}{l}{chessboard} & \multicolumn{1}{l}{fullscale} & \multicolumn{1}{l}{chessboard} \\
          \midrule
    1RC   & 98.52\% & 99.42\% & 86.69\% & {\bf 91.67\%} \\
    5RC   & 98.94\% & 97.71\% & 87.49\% & {\bf 89.14\%} \\
    \bottomrule
    \end{tabular}%
  \label{tab:nmnist}%
}
\end{table}

\subsubsection{JAFFE}
\label{sec:jaffe}
Since the size of the JAFFE database is limited, we use two methods to enhance the validation. 
First, we use 10-fold cross-validation technique. Second, we generate desired number of spike train samples by repeating the limited images.


Table~\ref{tab:jaffeacc} shows the inference accuracy of LSM on JAFFE using fullscale, scanline and chessboard pattern. 
The last line is the average accuracy derived from the ten-fold cross validation. It can be seen that chessboard pattern is still competitive with fullscale pattern, with about $\pm$3\% in accuracy. However, scanline performs much worse than the fullscale pattern on JAFFE2, with up to 7\% decrease in accuracy. The number of pixels from scanline pattern is similar to that from chessboard. Therefore, it could take several runs for randomly generating a set of scanlines that contain proper pixels and yield good performance. In other words, the performance will be worse if the scanlines are not properly chosen.

\begin{table}
    \centering
    \caption{Testing Accuracy Comparison on JAFFE.}
    \footnotesize{
      \begin{tabular}{p{1.565em}rrrrrr}
        \toprule
            & \multicolumn{3}{c}{JAFFE1} & \multicolumn{3}{c}{JAFFE2} \\
      No. & \multicolumn{1}{c}{FS} & \multicolumn{1}{c}{SL} & \multicolumn{1}{c}{{\bf CB}} & \multicolumn{1}{c}{FS} & \multicolumn{1}{c}{SL} & \multicolumn{1}{c}{{\bf CB}} \\
      \midrule
      \multicolumn{1}{c}{1} & 47.00\% & 43.00\% & 49.00\% & 62.00\% & 58.00\% & 61.00\% \\
      \multicolumn{1}{c}{2} & 43.00\% & 39.00\% & 50.00\% & 54.00\% & 53.00\% & 53.00\% \\
      \multicolumn{1}{c}{3} & 50.00\% & 55.00\% & 51.00\% & 64.00\% & 57.00\% & 58.00\% \\
      \multicolumn{1}{c}{4} & 29.00\% & 21.00\% & 38.00\% & 76.00\% & 56.00\% & 69.00\% \\
      \multicolumn{1}{c}{5} & 27.00\% & 27.00\% & 22.00\% & 54.00\% & 54.00\% & 53.00\% \\
      \multicolumn{1}{c}{6} & 29.00\% & 32.00\% & 31.00\% & 68.00\% & 57.00\% & 67.00\% \\
      \multicolumn{1}{c}{7} & 48.00\% & 45.00\% & 53.00\% & 64.00\% & 53.00\% & 52.00\% \\
      \multicolumn{1}{c}{8} & 48.00\% & 51.00\% & 44.00\% & 60.00\% & 60.00\% & 65.00\% \\
      \multicolumn{1}{c}{9} & 54.00\% & 51.00\% & 55.00\% & 66.00\% & 53.00\% & 64.00\% \\
      \multicolumn{1}{c}{10} & 42.00\% & 55.00\% & 49.00\% & 63.00\% & 58.00\% & 59.00\% \\
      \midrule
      \bf{AVG}   & 41.70\% & 41.90\% & \bf{44.20\%} & 63.10\% & 55.90\% & \bf{60.10\%} \\
      \bottomrule
      \end{tabular}%
      \\\vspace{1mm}\parbox{8.3cm}{FS means fullscale, SL means scanline, and CB means chessboard.}
      }
    \label{tab:jaffeacc}%
  \end{table}%

\subsection{Storage}
For storage, we refer to the input storage, i.e., spike records either generated from frame-based dataset or sampled from event-based dataset for training and testing, rather than the liquid state vector or the connection weights.
The reason for storing input spike records is that we want to feed the whole training samples into SNN framework at once rather than generating the spike trains for each sample one by one at simulation. The one-time feeding method greatly improve the learning efficiency and reduce the whole processing time \cite{liu2018event}.
The reduction in input storage by other input patterns compared with the fullscale pattern is obvious. With the same simulation time length per record, the reduction is proportional to the decrease in the number of input pixels. For instance, the number of pixels from chessboard pattern is only a quarter of that from fullscale pattern. So is the size of the spike records for training and testing, which is a 75\% reduction in input storage.


\subsection{Time}
\begin{figure}
  \includegraphics[width=.8\linewidth]{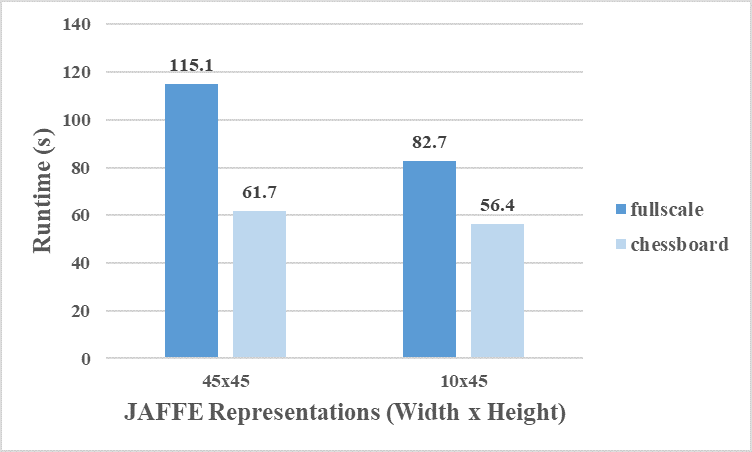}
  \caption{Average Runtime comparison on JAFFE.}
  \label{fig:jaffetime}
\end{figure}
Figure~\ref{fig:jaffetime} shows the runtime cost of using fullscale and chessboard pattern on JAFFE1 and JAFFE2. The figure is derived from averaging the result of ten simulations.
It is not as expected to reduce as much as the input storage. But nealy 30\% to 50\% reduction in runtime is also impressive. We find similar data in other experiments.
\section{Conclusion}
Liquid state machine (LSM) gets increasing popularity for it has less difficulty in training than other spiking neural networks. There are many efforts for enhancing LSM performance. For example, many works focus on exploring topology variety of LSM or enlarging the number of neurons in the liquid. These incurs computing and/or storage overhead. Large input space of LSM is also a factor that influence the LSM performance. We explore the influence of the reduction in input scale on LSM performance in this work via different input patterns. Using several input patterns and benchmark datasets, we show that input size reduction according to a certain input pattern can indeed have positive influence on the LSM performance with respect to accuracy, storage cost, and runtime. The best input pattern we discover is the chessboard pattern that can achieve nearly the same or better accuracy (2\% increase) at most occasions on frame-based datasets as well as the event-based dataset. At the same time, it can achieve up to 75\% input storage reduction and consumes only 50\% runtime in the best case. From the observations, we conclude that input size reduction is a good technique for saving computing and storage resource while keeping the same accuracy for real-world applications of LSM. And it can be integrated with variations of LSM architecture like the ensembles of small liquids. In the future, we will further explore how LSM will react to changes in input, like more complex input pattern, and validate it on more temporal datasets.

\bibliographystyle{ACM-Reference-Format}
\bibliography{jcst-lsm}

%
%
%
%
%
%
%
%

\end{document}